\documentclass{article} 
\usepackage[final]{colm2025_conference}

\usepackage{microtype}
\usepackage{hyperref}
\usepackage{url}
\usepackage{booktabs}
\usepackage{graphicx}
\usepackage{wrapfig}
\usepackage{multirow}
\usepackage{multicol}
\usepackage{lineno}
\usepackage{makecell}
\usepackage{tabularx}
\usepackage{array}
\usepackage{tcolorbox}
\tcbuselibrary{listingsutf8}
\usepackage{listings}
\usepackage{xcolor}

\definecolor{lightgray}{gray}{0.95}
\definecolor{darkblue}{rgb}{0, 0, 0.5}
\hypersetup{colorlinks=true, citecolor=darkblue, linkcolor=darkblue, urlcolor=darkblue}

\title{Inducing Programmatic Skills for Agentic Tasks}

\author{Zora Zhiruo Wang \quad Apurva Gandhi \quad 
{\bf  Graham Neubig} \quad {\bf Daniel Fried} \\
Carnegie Mellon University \\
{\tt \{zhiruow,apurvag,gneubig,dfried\}@cs.cmu.edu}}

\usepackage{xspace}
\newcommand{\awm}{\textsc{AWM}\xspace}
\newcommand{\asi}{\textsc{ASI}\xspace}

\usepackage{color-edits}
\addauthor{df}{cyan}
\addauthor{gn}{magenta}
\addauthor{zw}{orange}
\addauthor{ag}{brown}

\begin{document}

\ifcolmsubmission
\linenumbers
\fi

\maketitle
\vspace{-4mm}
\begin{abstract}
To succeed in common digital tasks such as web navigation, agents must carry out a variety of specialized tasks such as {\it searching for products} or {\it planning a travel route}.
To tackle these tasks, agents can bootstrap themselves by learning task-specific skills online through interaction with the web environment.
In this work, we demonstrate that \textit{programs} are an effective representation for skills. We propose \underline{a}gent \underline{s}kill \underline{i}nduction (\asi), which allows agents to adapt themselves by inducing, verifying, and utilizing program-based skills on the fly.
We start with an evaluation on the WebArena agent benchmark and show that \asi outperforms the static baseline agent and its text-skill counterpart by 23.5\% and 11.3\% in success rate, mainly thanks to the programmatic verification guarantee during the induction phase. \asi also improves efficiency by reducing 10.7--15.3\% of the steps over baselines, by composing primitive actions (e.g., \texttt{click}) into higher-level skills (e.g., \texttt{search\_product}).
We then highlight the efficacy of \asi in remaining efficient and accurate under scaled-up web activities. 
Finally, we examine the generalizability of induced skills when transferring between websites, and find that \asi can effectively reuse common skills, while also updating incompatible skills to versatile website changes.\footnote{\url{https://github.com/zorazrw/agent-skill-induction}}
\end{abstract}

\section{Introduction}
\vspace{-1mm}
To achieve success in common digital tasks such as web navigation, it is essential for agents to be able to perform a variety of specialized tasks such as searching for products on a shopping website \citep{yao2022webshop,deng2024mind2web} or finding a driving route on the map \citep{zhou2024webarena,xie2024travelplanner}.
While one source for agents to learn such tasks is demonstrations annotated by humans \citep{deng2024mind2web} or synthesized with large language models (LMs) on websites of interest \citep{murty2024bagel,murty2024nnetscape}, this can be a challenging offline learning procedure given the broad range of website domains and functionalities, especially for the collected demonstrations to match or cover the distribution of tasks queried at inference time \citep{zhou2024proposer}; not to mention the limitations in resources to collect abundant high-quality data at ease \citep{pan2024webcanvas}.

Instead of learning from demonstrations offline, an alternative way is to learn these tasks directly online from test queries to prevent potential distribution mismatch between demonstration and downstream tasks \citep{levine2020offline}. Some works propose to have agents induce casual abstractions \citep{majumder2024clin}, single-state guidelines \citep{fu2024autoguide}, or multi-step procedural workflows \citep{sarch2024vlm,wang2024agent} as a form of intermediate knowledge to augment agent memory via non-parametric approaches \citep{brown2020language}.
Nonetheless, most existing approaches represent this knowledge in text, offering limited quality and verification guarantees.
In this work, we propose that \emph{executable programs} are effective representations for intermediate skill acquisition, given their verifiability and composability advantages \citep{setlur2025scaling}.

\begin{wrapfigure}[20]{r}{0.55\textwidth}
\vspace{-4mm}    
\includegraphics[width=0.54\textwidth]{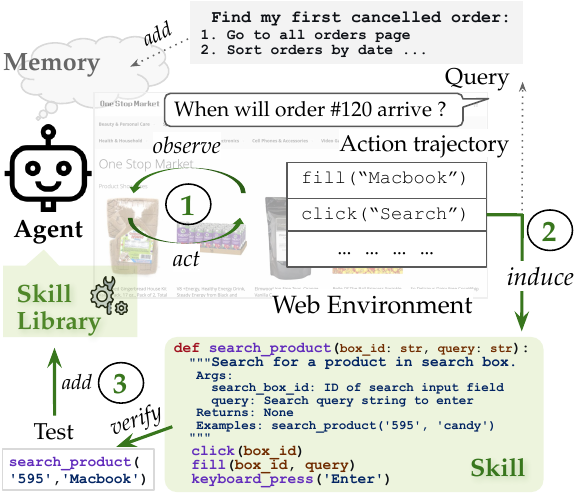}
\vspace{-2mm}
\caption{Online adaptive agent that induces and reuses programmatic skills as actions (bottom), as opposed to adding textual skills in memory (top).}
\label{fig:pipeline}
\end{wrapfigure}

We present \asi, namely \underline{a}gent \underline{s}kill \underline{i}nduction (\S\ref{sec:2:method-asi}), that induces and applies programmatic skills along the process of solving user web navigation queries. More concretely, given a natural language (NL) query, the agent first generates an action trajectory attempting to solve the task using built-in, primitive actions such as \texttt{click} and \texttt{scroll}. The agent then induces higher-level skills (e.g., \texttt{search\_product(name)}) that wrap primitive actions or prior skills as executable programs, accompanied with corresponding test trajectories to verify their quality. Verified skills are then incorporated into the agent action space and can be directly called to solve future tasks with similar procedures, as depicted in \autoref{fig:pipeline} (bottom).

We first evaluate \asi on the WebArena benchmark \citep{zhou2024webarena} (\S\ref{sec:expr-basic}) and demonstrate that our online, adaptive \asi surpasses its static agent baseline without adaptive components by 23.5\% in success rate. To validate the advantage of using programmatic representations for skills, we further compare to an adaptive agent, \awm \citep{wang2024agent}, that represents skills in memory as non-executable texts (\autoref{fig:pipeline} top); we find \asi scores 11.3\% higher success rate by employing verifiable, programmatic skills (\autoref{fig:pipeline} bottom). Beyond the correctness aspect, the task-solving procedures by \asi-supported agents are 10.7--15.3\%  more efficient than the baseline approaches, mainly because of the action space abstraction and composition enabled by the programmatic skill representation.

We further stress test \asi on scaled-up scenarios (\S\ref{sec:expr-scaled-up}) that require substantially longer-horizon trajectories to complete the task. Across various domains such as shopping and social forums, we find the efficiency advantage offered by \asi is more pronounced, reducing action trajectories by 9.5 and 5.6 average steps compared to static and text-form agents. Facilitated by this improved procedural efficiency and planning, we find that \asi agent adheres to the optimal trajectory better and achieves tasks 20.7--38.9\% more correctly.

Finally, we study agent behaviors on generalizing induced skills to other websites (\S\ref{sec:expr-cross-web}), particularly from sandboxed, simulated websites to real-world websites of similar domains. 
While \asi agents effectively transfer common skills (e.g., \texttt{search\_product}) to new websites, some skills may be incompatible with alternative website designs hence less useful. Nonetheless, \asi can quickly refine its prior skills or create new skills on new websites, indicating it allows agents to adapt online while maintaining verifiability via programs.

In short, \asi enhances web agent success and efficiency by inducing and applying verifiable programmatic skills, in general and longer-horizon tasks, even across varied websites. 
\section{Agent Skill Induction}
\label{sec:2:method-asi}
\vspace{-1mm}

In this section, we first lay out the web agent problem setup (\S\ref{sec:2.1:problem}) and introduce online, self-adaptive agents  (\S\ref{sec:2.2:intro-induce}). We then describe the core component of \asi --- programmatic skill induction and verification (\S\ref{sec:2.3:skill-induce}).

\subsection{Problem Statement: Online Adaptive Agent}
\label{sec:2.1:problem}
\vspace{-1mm}

For the scope of this work, we focus on language model (LM) based agents, where each agent policy consists of an LM backbone $\mathcal{L}$, a memory $\mathcal{M}$, and a skill library $\mathcal{A}$, as illustrated in \autoref{fig:pipeline} top and bottom. In the implementation, the memory $\mathcal{M}$ and the skill library $\mathcal{A}$ are provided as input context to the LM backbone. We denote the agent policy as $\pi_{\mathcal{L}}(\cdot|\mathcal{M}, \mathcal{A})$ and $\pi_{\mathcal{L}}$ for short.
We focus on the web browser environment defined by a transition function $\mathcal{T}(s' | s, a)$ that models the change in the webpage after an action.

We focus on an online adaptation scenario where we have access to a sequence of NL queries $Q = \{q_1, q_2, \cdots, q_N\}$ specifying the tasks, and no other information such as demonstration trajectories or ground-truth rewards are available \citep{wang2024trove,wang2024agent}.
For each task specified by a natural language (NL) query $q$, the agent generates a trajectory of actions $\tau = (s_0, a_0, s_1, a_1, \cdots, s_{H-1}, a_{H-1}, s_H)$ for a finite number of $H$ steps. At each time step $h$ in the horizon, the agent receives observation $o_h$ from the current state $s_h$, and generates an action $a_h \in \mathcal{A}$ based on the observations and actions so far, via $\pi_{\mathcal{L}}(o_{0:h}, a_{0:h-1}; \mathcal{M}, \mathcal{A}) \rightarrow a_h$.
The generated action will be executed on the environment and incurs a state change $\mathcal{T}(s_h, a_h) \rightarrow s_{h+1}$. This observe--act loop continues for $H$ steps until the task reaches a task-terminating condition, such as the agent generating a termination action (e.g., \texttt{send\_msg\_to\_user}) or the horizon reaches a pre-determined maximum number of steps $h = H_{max}$. 
We denote each pair of query and trajectory $(q, \tau) := e$ as an episode $e$.
Agents can update the content in $\mathcal{M}$ and $\mathcal{A}$ and reuse them across episodes. 

\subsection{Inducing Reusable Skills}
\label{sec:2.2:intro-induce}
\vspace{-1mm}

To realize online adaptive agents, one common approach is to induce skills from correct trajectories to update the agent \citep{wang2024agent}. But since ground-truth rewards are unavailable, an LLM-based evaluator $V_{\mathcal{L}}$ is often used to judge the correctness of episodes. Formally, from the total of $N$ episodes throughout the online process $\{e^1, \cdots, e^N\} := \mathcal{E}$, we employ an LM-based evaluator $V_{\mathcal{L}}(e) \rightarrow 0/1$ to filter out the episodes predicted as correct $\mathcal{E}_V = \{e_i \in \mathcal{E}| V_{\mathcal{L}}(e_i)=1, i \in \{1, \cdots, N\}\}$ and perform skill induction only on $\mathcal{E}_V$.

Central to our adaptive agents is an induction component $I$ that enables the adaptivity of agents, which can be rule-based \citep{ellis2023dreamcoder,grand2024lilo} or instantiated by an LM $I(\cdot | LM)$ \citep{wang2024agent}; we follow the latter for its better performance and use $I$ to represent the module for simplicity. 
For online adaptive agents $\pi_{\mathcal{L}}$, to induce skills, 
$I$ is instructed to take in one filtered episode $e$ and output one or more pieces of desired skills $D = \{d\}$, denoted as $I(e) \rightarrow \mathcal{D}$. 
Following AWM~\citep{wang2024agent}, we update the agent in non-parametric ways that incorporate the induction outcome $I(e_t) \rightarrow d_t$ into the agent, instead of updating the parameters of the underlying LM backbone $\mathcal{L}$ for agent policy $\pi_{\mathcal{L}}$.

Unlike \awm which represents skills in free-form text representations and can only augment agent memory via $\mathcal{M}_t \cup \{d_t\} \rightarrow \mathcal{M}_{t+1}$ (\autoref{fig:pipeline} top), we introduce \asi that represents skills as executable python programs, and directly integrate skills into the agent action space instead, via $\mathcal{A}_t \cup \{d_t\} \rightarrow \mathcal{A}_{t+1}$ (\autoref{fig:pipeline} bottom).

\subsection{Inducing and Verifying Programmatic Skills}
\label{sec:2.3:skill-induce}
\vspace{-1mm}

To improve the induction quality, we propose a change in representation \textit{from free-form text to executable programs}, which offers advantages in correctness and efficiency. 
For one, the program format enables ready verification on skill correctness by executing them; for another, skill programs abstract multiple lower-level actions into a higher-level function call, thus agents can solve tasks in fewer steps without tackling tricky low-level details.

\noindent \textbf{Inducing Programmatic Skills} \quad
We first clean the input episodes to ensure the induction quality.
We remove all the steps that cause execution errors such as invalid argument format, to keep these invalid actions from distracting agent predictions. Furthermore, noticing the long and possibly redundant thought process generated by agents along with each action, we simplify each thought text paragraph into a short one-sentence description (e.g., ``Clicked the directions button to access the route planning feature'') using LM, effectively reducing the thought content from 87.9 to 13.4 tokens per step.

\begin{figure*}[t!]
\vspace{-4mm}
\centering
    \includegraphics[width=0.93\textwidth]{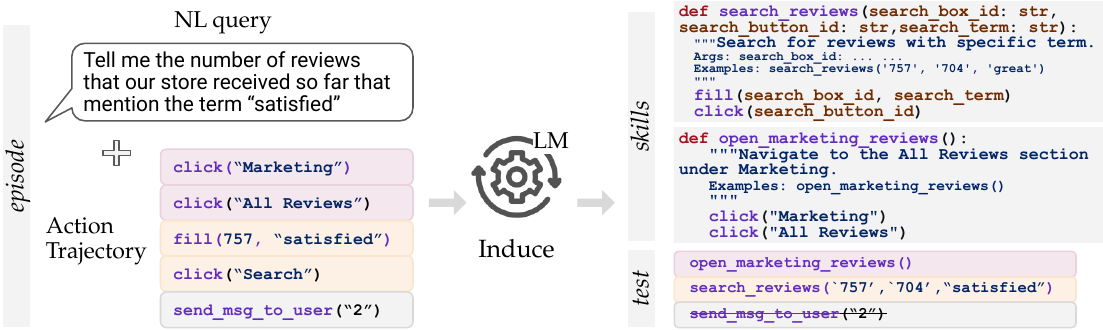}
\vspace{-2mm}
\caption{Inducing programmatic skills and rewriting the trajectory from an episode.}
\label{fig:skill-induction}
\vspace{-2mm}
\end{figure*}

Given a clean input episode $e$, we now prompt the induction module $I$ to produce one or more program functions to represent reusable skills $\mathcal{D} = \{d\}$ as executable programs.
As exemplified in \autoref{fig:skill-induction}, given the input episode on the left side, the induction module first produces two skills \texttt{open\_marketing\_reviews()} and \texttt{search\_reviews(search\_box\_id, search\_button\_id, search\_term)} in the form of callable program functions. 

\noindent \textbf{Skill Verification} \quad
With the programmatic nature of \asi's skills, we can readily verify their correctness by executing them and checking if tasks can be solved successfully. 
While a naive way is to query the agent with the same NL query and allow it to use newly induced skill actions, we find agents may not always use new skills due to the large search space of possible action trajectories. To have agents more efficiently generate trajectories that test skills in a more targeted way, we curate a rewritten trajectory prefix $\tau_{D}$ to constrain the first few steps executed in the environment, by rewriting and truncating the input action trajectory $\tau$, and subsequently asking the agent to complete the prefix to get a full, checkable trajectory $\tau_f$.
Concretely, we first take the original action trajectory in the input episode $\tau$ (consisting of primitive actions or previously learned skills), and ask the induction module $I$ to transform it to a skill-using trajectory (\autoref{fig:skill-induction} bottom right), by replacing sub-trajectories in $\tau$ with calls to the newly induced skill programs $\mathcal{D}$, if possible. Zooming into the \autoref{fig:skill-induction} example, this procedure merges \texttt{click(`Marketing')} $\rightarrow$ \texttt{click(`All Reviews')} to an \texttt{open\_marketing\_reviews()} call; transforms \texttt{fill(757, `satisfied')} $\rightarrow$ \texttt{click(`Search')} to a call of the second skill \texttt{search\_reviews(`satisfied')} with the specified term `satisfied'; and adopted the last \texttt{send\_msg\_to\_user(`2')} step directly. 
Note that we follow \citet{wang2024agent} and induce skills according to each website, so some skills could be tailored to particular webpage contexts such as the `Marketing' and `All Reviews' link constants in \texttt{open\_marketing\_reviews}, while other skills apply to more versatile setups such as searching for different reviews in \texttt{search\_reviews}.

Next, to avoid spurious successes in skill verification, we truncate the trajectory yielded above by removing any trailing primitive actions after the last call to a skill program.
Taking \autoref{fig:skill-induction} as an example, in the original input trajectory, the last \texttt{send\_msg\_to\_user(`2')} already sends over the correct answer \texttt{`2'} to the user. However, if we directly adopt this last step into the skill-using trajectory $\tau_D$, then executing it will always return the correct message to the user, regardless of whether the previous skill calls are valid. 
We thus remove such trailing actions to make sure verification attends to the induced skills we are testing.

After rewriting and truncation, we get this skill-using trajectory $\tau_D$ as the prefix to test skills. We now query the agent $\pi$ again with the same NL query $q$ and first execute $\tau_{D}$ on the environment. We then allow agents to continue generating up to $H_{max} - |\tau_D|$ actions to finish the task.
In the \autoref{fig:skill-induction} example, to successfully solve query $q$, we expect the agent to generate another step of \texttt{send\_msg\_to\_user(...)} with the correct answer \texttt{`2'} in the message.
We take the concatenation of the trajectory prefix $\tau_D$ and the later additionally produced new steps $\tau_A$ (e.g., $[$\texttt{send\_msg\_to\_user(`2')}$]$) as the full trajectory $\tau_{f}$. We then decide whether to add the induced programs $\mathcal{D}$ into the agent skill library as $\mathcal{A}$ by examining $\tau_{f}$.

Specifically, we check $\tau_{f}$ from three dimensions:
(1) \textit{Correctness}: if executing $\tau_{f}$ successfully solves the task $q$ as judged by the neural model evaluator $V_{\mathcal{L}}$; 
(2) \textit{Skill Usage}: if the trajectory contains at least one call to at least one new skill in $\mathcal{D}$; and 
(3) \textit{Skill Validity}: if all skill-calling actions cause environment changes.
If all three boxes are checked, we add the skills being called in the trajectory $\tau_{f}$ to the agent skill library $\mathcal{A}_t \cup \mathcal{D}_{\it called} \rightarrow \mathcal{A}_{t+1}$. By adding $\mathcal{D}_{\it called}$, the agent can now generate actions that call these skill programs to solve subsequent tasks.

\section{General Web Navigation Performance}
\label{sec:expr-basic}

\subsection{Experiment Setup}

\noindent \textbf{Benchmark and Evaluation} \quad
To evaluate \asi on general web navigation scenarios, we adopt the WebArena benchmark \citep{zhou2024webarena} that contains 812 test examples covering five major web activity domains: e-commerce, social forum, software development, content management, and travel.
Each example in WebArena has an NL query $q$ for the task, and a program-based evaluator that provides a binary 0/1 score for any given trajectory $\tau$ to judge if it successfully solves the task $q$. This program-based evaluator enables relatively rigorous evaluation based on the functional correctness of the action trajectory. We report the average score across all WebArena examples, if not specified otherwise.

\noindent \textbf{Backbone LM and Agent Architecture} \quad
We use the top-performing \texttt{claude-3.5-sonnet} model as the LM backbone for all components, including the agent policy $\pi$, the neural evaluator $V$, and the skill induction modules $I$.
For experimentation, we use the BrowserGym \citep{chezelles2024browsergym} framework, which takes the webpage accessibility tree as observation, and instantiates the skill library $\mathcal{A}$ with the WebArena default action space listed in \S\ref{app:expr-details}.

\noindent \textbf{Baselines} \quad
We take the vanilla Claude model with the BrowserGym framework \citep{drouin2024workarena} as the non-adaptive agent baseline. Additionally, we compare \asi to \awm \citep{wang2024agent}, the current top-performing online adaptive web agent method. Because \awm was originally developed with the \texttt{gpt-4o} model, for a fairer comparison, we also experiment with \awm with \texttt{claude-3.5-sonnet} model as its LM backbone and also apply the episode cleaning procedure to enhance induction quality.
We compare the two baseline methods with our \asi approach. We provide the complete prompts for each agent component: task-solving, episode evaluation, episode cleaning, and skill induction, in \S\ref{app:expr-details}.

\subsection{Results and Analysis}

\begin{table}[t!]
\centering
\small
\vspace{-3mm}
\resizebox{0.99\textwidth}{!}{
\begin{tabular}{ll|cc|cccccc}
\toprule
\multicolumn{1}{c}{\bf Model} & \multicolumn{1}{c|}{\bf Method} & {\bf \# Steps} & {\bf SR} & {Shop} & {Admin} & {Reddit} & {GitLab} & {Maps} & {Multi} \\
\midrule
\multirow{2}{*}{GPT} & {Vanilla} & {-} & {12.3} & {13.9} & {10.4} & {$~~$6.6} & {15.0} & {15.6} & {8.3} \\
{} & {\awm} & {5.9} & {35.5} & {32.1} & {29.1} & {\bf 54.7} & {\bf 35.0} & {42.2} & {18.8} \\
\midrule
\multirow{3}{*}{Claude} & {Vanilla} & {5.6} & {32.7} & {32.6} & {36.8} & {36.8} & {26.1} & {38.5} & {\bf 20.8} \\
{} & {\awm} & {5.9} & {36.3} & {34.8} & {39.0} & {51.9} & {28.9} & {39.4} & {18.8} \\
{} & {\bf \asi (ours)} & {\bf 5.0} & {\bf 40.4} & {\bf 40.1} & {\bf 44.0} & {\bf 54.7} & {32.2} & {\bf 43.1} & {\bf 20.8} \\
\bottomrule
\end{tabular}
}
\vspace{-1mm}
\caption{WebArena success rate by adaptive agents with programmatic skills, in comparison to a static vanilla agent baseline, and a text-skill learning adaptive agent.}
\label{tab:webarena-results}
\vspace{-3mm}
\end{table}

In \autoref{tab:webarena-results}, compared to the vanilla static-agent baseline, adaptive agents (\awm and \asi) generally achieve 11.0--23.5\%  higher success rates overall.
Among adaptive agents, our \asi with programmatic skills, achieves another 11.3\%  success rate gain across websites, compared to its \awm counterpart that induces and uses textual skills.
Meanwhile, \asi offers additional efficiency benefits by reducing the number of steps in solutions by 15.3\% and 10.6\% than vanilla and \awm agents, as one skill-call action can often execute multiple steps written in primitive actions used by vanilla and \awm agents. 
These advantages in correctness and efficiency are exhibited prominently across different websites and tasks, as shown by the website breakdown on \autoref{tab:webarena-results} (right). Refer to \S\ref{app:skill-analysis} for more analysis.

\subsection{Why are Programmatic Skills Better?}
To more concretely answer why programmatic skills are more effective than textual skills, we take a closer look on the two main differences between \awm and \asi: [1] whether the induction outcome is verified via execution, and [2] whether the induced skills are provided in memory for reference purpose only, or in the action space that allows execution.

\noindent \textbf{Better Induction Quality} \quad
We take the shopping website as a representative, and analyze the textual and program skills induced by \awm and \asi agents. 
We group textual and program skills by their functionality and show one representative example in \autoref{tab:induce-ablation}.
Compared to the clear functional boundary and highly-reusable granularity of the \texttt{search\_product} skill, we find that the textual skills often have (1) more redundant steps, (2) example-specific context: e.g., the last text skill aims to find `game accessories' while the steps generally applies to any product, and (3) fuzzier boundaries between separable tasks, e.g., the first skill mixes \texttt{product-search} and \texttt{add-to-wishlist} procedures, thus may not offer optimal guidance when asked to, e.g., search product and add it to cart instead.

\begin{table}[ht]
\vspace{-3mm}
\centering
\small
\begin{tabular}{cc}
\toprule
{\bf Programmatic Skills} & {\bf Textual Skills} \\
\midrule
\multirow{6}{*}{\includegraphics[width=0.49\textwidth]{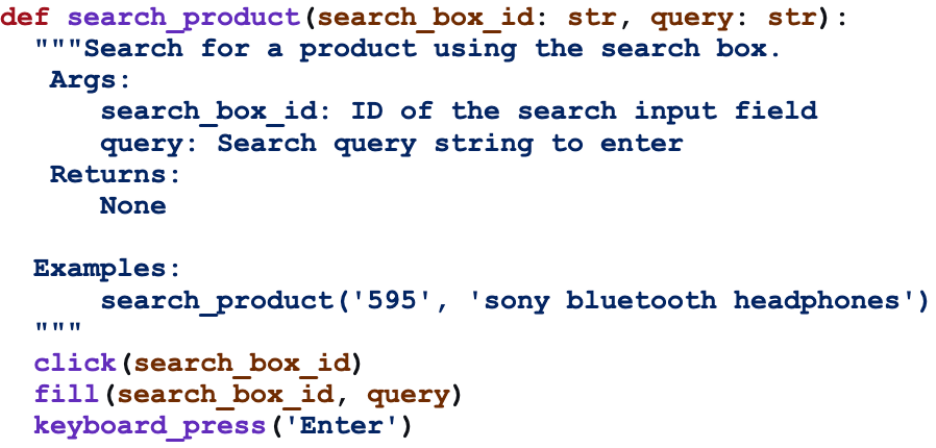}} & {\includegraphics[width=0.44\textwidth]{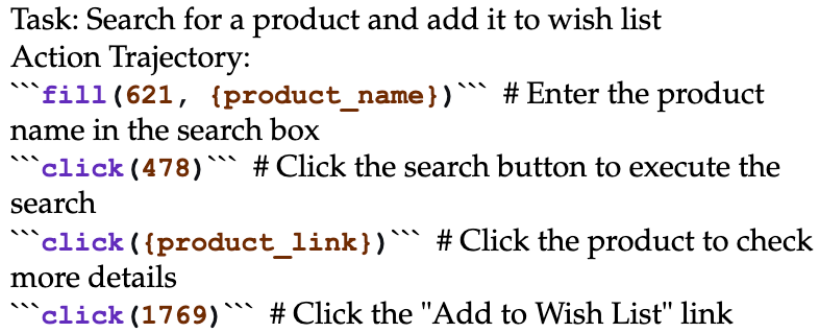}} \\
\cmidrule{2-2}
{} & {\includegraphics[width=0.44\textwidth]{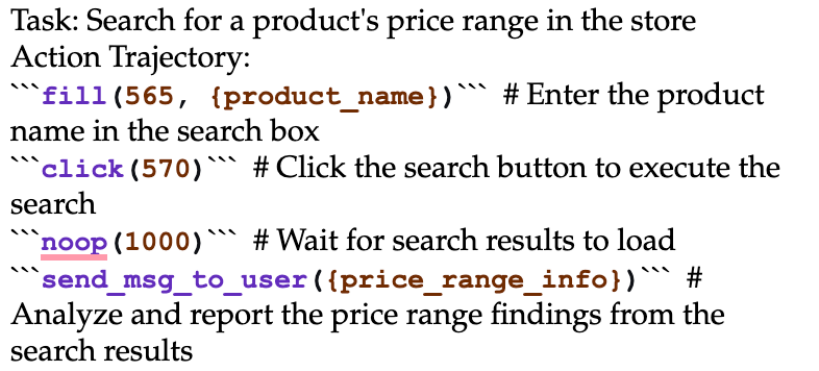}} \\
\cmidrule{2-2}
{} & {\includegraphics[width=0.44\textwidth]{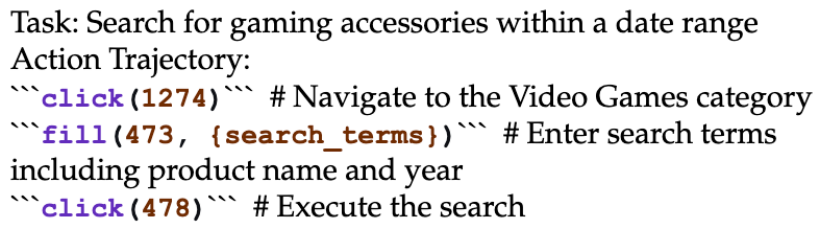}} \\
\bottomrule
\end{tabular}
\vspace{-1mm}
\caption{Example textual and program skills induced on the shopping website.}
\vspace{-2mm}
\label{tab:induce-ablation}
\end{table}

\noindent \textbf{Verified Induction Improves End Success Rate} \quad
From qualitative examination of the induction outcomes, we find roughly similar numbers of episodes evaluated as correct and used for induction (70 and 58 examples for \awm and \asi), \asi produced programs pass verification for only 15.6\% of the turns, whereas \awm adds new skills for 31.4\% of the time (replace or add none otherwise).
While skill usage (in memory or as action, [2]) is designated for \awm and \asi, we hypothesize that verification [1] affects induction quality and thus end success. 
We thus experiment with another setting that induces programs (such that verification is enabled), and only use the induced skills in memory, to study the importance of induction quality.
As shown in \autoref{tab:format-ablation}, inducing skills with execution-based verification (i.e., {\it {\it (unverified, text)} $\rightarrow$ (verified, program)}), while always present skills in memory, improves end success rate by 4.2 points, indicating the importance of higher-quality induction via verification. 
Yet it is still 3.7 points lower than \asi, suggesting the incompatibility of program format to agent memory. Indeed, we observe many cases where the agent tries to call the skill programs but unsuccessfully, since they are not supported in the action space.

\begin{wraptable}[10]{r}{0.50\textwidth}
\centering
\small
\vspace{-3mm}
\resizebox{0.50\textwidth}{!}{
\begin{tabular}{llc}
\toprule
\multicolumn{2}{c}{\bf Method} & {\bf SR} \\
\midrule
\multirow{3}{*}{Add to Memory} & {unverified, text} & {32.6} \\
{} & {verified, program} & {36.4} \\
{} & {verified, text} & {39.0} \\
\midrule
{Add as Actions} & {verified, program} & {40.1} \\
\bottomrule
\end{tabular}
}
\vspace{-3mm}
\caption{Ablation study on induction verification and format on the shopping website.}
\label{tab:format-ablation}
\end{wraptable}

\noindent \textbf{Textual Representations Suit Memory Better} \quad
To prevent the agent from trying to call these plausible programs, we ablate another setting that transforms program skills to textual format (as \autoref{tab:induce-ablation} right) and provide them in agent memory, dubbed {\it (verified, text)}.
This format transformation effectively improves the overall success rate by another 2.6 points, getting a little closer to \asi.
Given the different downstream usage, i.e., memory or actuation, textual and program formats may suit individual scenarios better.

Beyond basic web navigation tasks, in the next two sections, we examine agents in two other important scenarios, scaled-up activities (\S\ref{sec:expr-scaled-up}) and cross-website generalization (\S\ref{sec:expr-cross-web}).

\section{Scaled-Up Browsing Activities}
\label{sec:expr-scaled-up}

The WebArena benchmark mainly features isolated, single-task scenarios, such as adding a single product to the shopping cart. However, in real-world practices, people need to do a series of such tasks together, such as adding multiple related products (e.g., coffee and mug) to the cart before finally checking out. This browsing request can lead to extremely long-horizon tasks, sometimes with repetitive intermediate procedures.
We identify this to be a scenario to further demonstrate the efficacy of program skills, as opposed to textual skills, as programs lend themselves naturally to repeated invocation and composition.

\begin{figure*}[ht]
\vspace{-2mm}
\centering
    \includegraphics[width=1.0\textwidth]{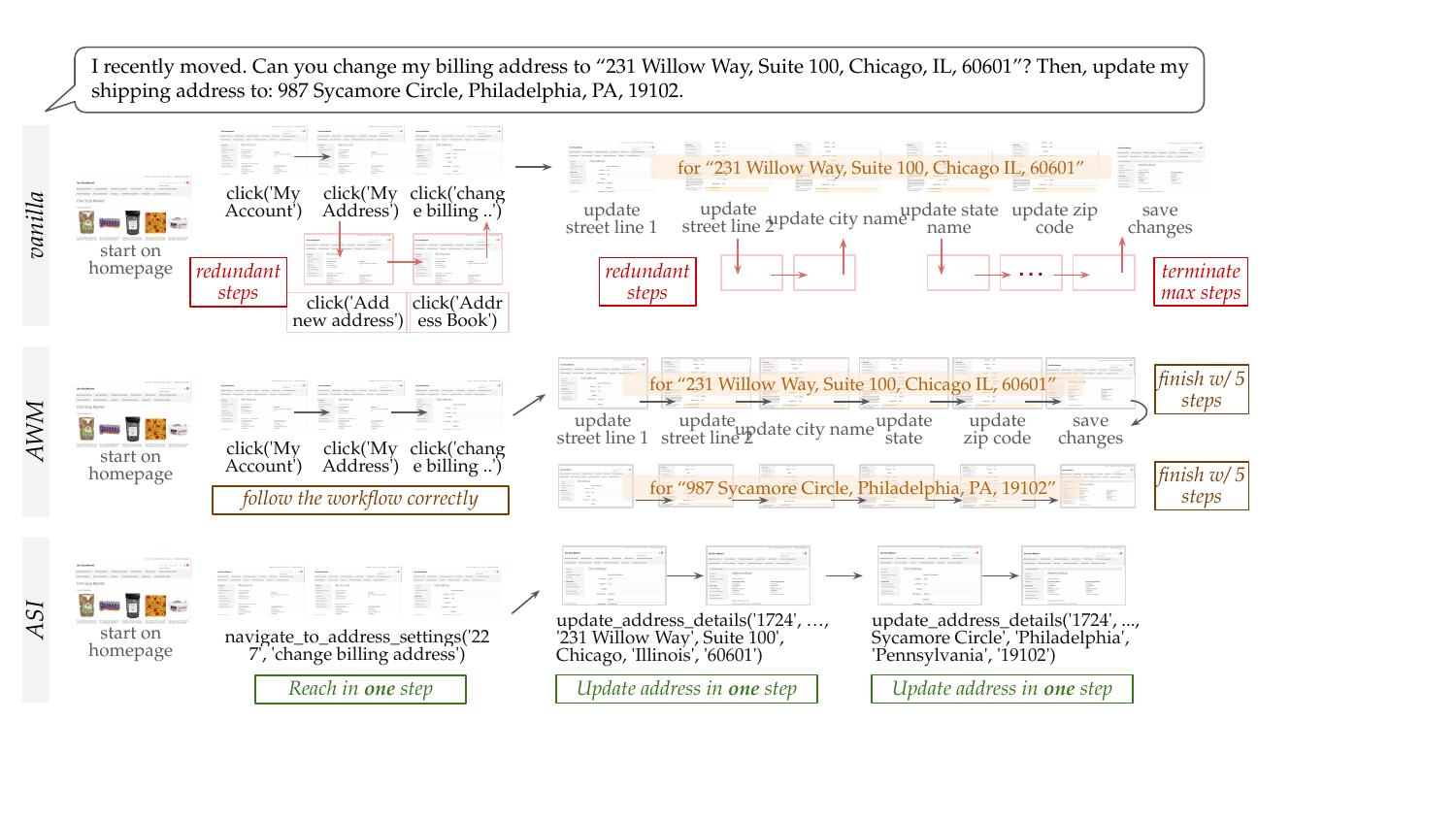}
\vspace{-5mm}
\caption{Example scaled-up task of updating multiple addresses on shopping website.}
\label{fig:scaled-up-example}
\vspace{-1mm}
\end{figure*}

Therefore, we curate several case scenarios where the user asks for action-dense instructions, such as the tasks listed in \autoref{fig:scaled-up-example}. Because the tasks are long-horizon and involve multiple sub-tasks, we follow \citet{xu2024theagentcompany} and set up intermediate checkpoints to better track the intermediate progress of agents. Refer to \S\ref{app:b.1:scale-up-tasks} to see the full list of tasks and their evaluation checkpoints.
We measure the success rate of each example by the percentage of checkpoints achieved by the agent. We report the average success rate of all examples, as well as the average number of steps taken to solve the tasks, in \autoref{tab:scaled-up-results}.


\begin{table}[ht]
\centering
\small
\vspace{-1mm}
\resizebox{1.0\textwidth}{!}{
\begin{tabular}{l|cc|cc|cc|cc|cc}
\toprule
\multirow{2}{*}{\bf Method} & \multicolumn{2}{c|}{Shopping} & \multicolumn{2}{c|}{Admin} & \multicolumn{2}{c|}{Reddit} & \multicolumn{2}{c|}{GitLab} & \multicolumn{2}{c}{Map} \\
{} & {sr $\uparrow$} & {\# steps $\downarrow$} & {sr $\uparrow$} & {\# steps $\downarrow$} & {sr $\uparrow$} & {\# steps $\downarrow$} & {sr $\uparrow$} & {\# steps $\downarrow$} & {sr $\uparrow$} & {\# steps $\downarrow$} \\
\midrule
{\sc Vanilla} & {$~~$41.7} & {23.5} & {58.0} & {20.8} & {33.3} & {23.0} & {33.3} & {40.0} & {$~~$40.0} & {15.2} \\
{\awm} & {$~~$68.3} & {21.5} & {74.0} & {18.2} & {40.0} & {16.8} & {50.0} & {33.8} & {$~~$65.0} & {12.6} \\
{\asi (ours)} & {\bf 100.0} & {\bf 16.3} & {\bf 91.0} & {\bf 14.2} & {\bf 55.0} & {\bf 12.8} & {\bf 55.0} & {\bf 25.4} & {\bf 100.0} & {\bf $~~$6.2} \\
\bottomrule
\end{tabular}
}
\vspace{-2mm}
\caption{Performance of vanilla, \awm, and \asi agents in scaled-up browsing scenarios. We perform statistical testing between \asi and each baseline and verify all improvements are statistically significant with t-statistics $|t|>2$ and $p<0.05$; see \S\ref{app:b.3:significance-testing} for more details.}
\label{tab:scaled-up-results}
\end{table}

\noindent \textbf{\asi Features Improved Efficiency} \quad
Across all websites, \asi-produced trajectories have 6.6--14.6 and 4.0--8.4\% fewer steps, compared to vanilla and \awm baselines, respectively. As the task horizon continues to grow when involving more intermediate checkpoints, this margin between \asi and baselines will predictably be more prominent.

\noindent \textbf{Subsequent Benefits in Success Rate} \quad
\asi also achieves higher success rates with more efficient trajectories, outperforming vanilla and \awm baselines by 38.9\% and 20.7\% on average. From manual analysis, we find this improvement comes from easier, better agent planning when using higher-level skills, without the need to tackle more complex procedures if only low-level primitive actions are available, as with vanilla and \awm agents.

\noindent \textbf{Case Study: Changing Multiple Addresses} \quad
We present a representative case on the shopping website: changing billing and shipping addresses after moving. 
As depicted in the top row in \autoref{fig:scaled-up-example}, the vanilla agent without adaptive skills often roams into some irrelevant exploration steps, instead of sticking to the optimal route to solve the required task. It runs for minutes and exhausts the maximum steps (i.e., 50) before finishing the task.

With \awm, adding textual skills in memory provides soft guidelines for agents to follow, the agent thus better sticks to the goal and finishes each part of the task (i.e., navigate to the address page, update billing/shipping address) step by step. Although successful, the trajectory is long, i.e., 27 steps, and still takes a few minutes to finish.

In comparison, \asi (in \autoref{fig:scaled-up-example} bottom row) showcases its efficiency by using learned skills to \texttt{navigate\_to\_address\_settings} and \texttt{update\_address\_details} can solve each part in one step (vs. the 3--6 steps used by \awm for these parts). Overall, \asi correctly finishes all required actions in only 4 steps, shortening the horizon by 85.2\% compared to \awm.
\section{Adapting Across Websites}
\label{sec:expr-cross-web}

\begin{wraptable}[8]{r}{0.60\textwidth}
\vspace{-4mm}
\centering
\small
\resizebox{0.59\textwidth}{!}{
\begin{tabular}{l|cc}
\toprule
\multicolumn{1}{c|}{\bf Domain} & {\bf WebArena Sandboxed} & {\bf Real-World} \\
\midrule
{shopping} & {OneStopMarket} & {Target} \\
{online forum} & {PostMill} & {Reddit} \\
{travel} & {OpenStreetMap} & {Google Maps} \\
\bottomrule
\end{tabular}
}
\vspace{-2mm}
\caption{Real-world in-domain website counterparts to each WebArena sandboxed website.}
\label{tab:website-selection}
\vspace{-1mm}
\end{wraptable}

To examine whether agents can generalize with learned skills, we test agents on real-world website counterparts for some of the domains in WebArena as listed in \autoref{tab:website-selection}.
\footnote{We did not test on administrative and software websites given their more severe safety concerns.}
This experiment setup can reflect on (1) transfer across different websites of the same domain, and (2) transfer from simulated, sandboxed to real-world websites.

For each sandbox-real website pair, we take ten information-seeking style queries \citep{he2024webvoyager} in WebArena that do not involve potential privacy leakage or unrecoverable risky actions, such as making a purchase or changing user password. We provide the task details in \S\ref{app:b.2:cross-web-tasks}.
We compare \asi and \awm with their programmatic and textual skills as learned in \S\ref{sec:expr-basic}, as well as comparing to the vanilla static agent baseline.

\begin{wrapfigure}[15]{r}{0.65\textwidth}
\vspace{-3mm}    
\includegraphics[width=0.64\textwidth]{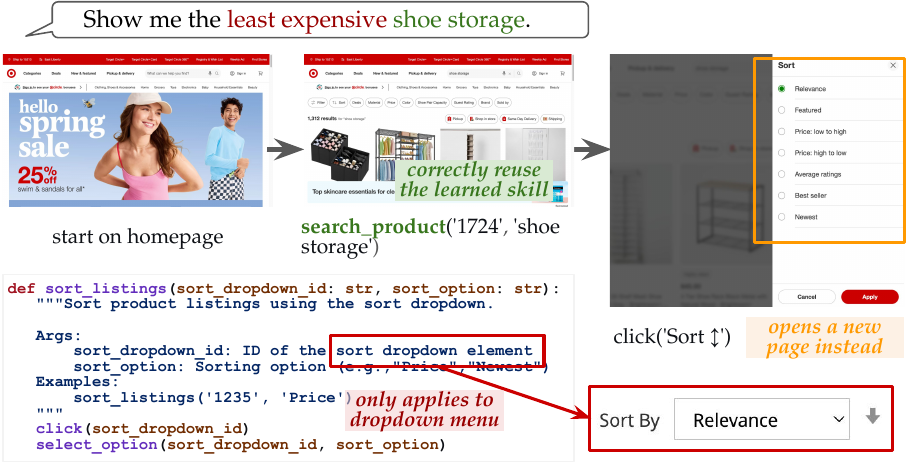}
\vspace{-2mm}
\caption{\asi can generalize the \texttt{search\_product} skill but face incompatibility when sorting items.}
\label{fig:crossweb-example}
\end{wrapfigure}

\noindent \textbf{Transferring Common Skills} \quad
In \autoref{fig:crossweb-example}, we can see how \asi can effectively reuse common skills such as \texttt{search\_product} in the first step on the Target website.

\noindent \textbf{Incompatible Skills} \quad
One challenge faced by \asi is that some prior skills become incompatible on the new website. For example, the \texttt{sort\_by\_listings()} induced on OneStopMarket selects options from a dropdown menu, yet sorting on the Target website opens a sidebar; despite their semantic similarity, the concrete actions in skill programs are no longer applicable. Still, we find that agents can often spot this incompatibility and rarely attempt to use these deprecated skills.

\noindent \textbf{Adapting Skills to New Environment} \quad
Although some skills induced on previous websites 
\begin{wraptable}[9]{r}{0.68\textwidth}
\centering
\small
\vspace{-3mm}
\resizebox{0.67\textwidth}{!}{
\begin{tabular}{l|cc|cc|cc}
\toprule
\multicolumn{1}{c|}{\multirow{2}{*}{\bf Method}} & \multicolumn{2}{c|}{\bf Shopping}  & \multicolumn{2}{c|}{\bf Reddit} & \multicolumn{2}{c}{\bf Map} \\
{} & {sr $\uparrow$} & {\# steps $\downarrow$} & {sr $\uparrow$} & {\# steps $\downarrow$} & {sr $\uparrow$} & {\# steps $\downarrow$} \\
\midrule
{Vanilla} & {80.0} & {5.4} & {40.0} & {4.8} & {$~~$63.3} & {7.4} \\
{\awm} & {80.0} & {5.0} & {56.7} & {4.8} & {\bf 100.0} & {6.2} \\
{\asi} & {\bf 90.0} & {3.4} & {\bf 76.7} & {4.4} & {$~~$93.3} & {4.4} \\
\midrule
{\awm + update} & {80.0} & {5.4} & {63.3} & {5.8} & {\bf 100.0} & {7.2} \\
{\asi + update} & {\bf 90.0} & {\bf 3.2} & {\bf 76.7} & {\bf 4.0} & {$~~$93.3} & {\bf 4.2} \\
\bottomrule
\end{tabular}
}
\vspace{-2mm}
\caption{\footnotesize Cross-website results. \asi significantly surpasses baselines in {\it sr} and {\it \# steps} (with $|t|>2$ and $p<0.05$) from our analysis in \S\ref{app:b.3:significance-testing}.}
\label{tab:crossweb-results}
\vspace{2mm}
\end{wraptable}
cannot be directly used on arbitrary new websites, we hypothesize that these skills can still serve as informative references on solving procedurally similar tasks or composing new skills targeted for the new website design.

We thus allow agents to induce new skills or update previously acquired skills from experiences on the new website, denoted as {\it +update} entries in \autoref{tab:crossweb-results}. 
We find that enabling skill update in both textual and program formats helps agent performance on new websites. Within the short online learning process (tens of examples), \awm adapts faster to the new websites, while \asi sees a more pronounced improvement in efficiency.

\section{Related Work}

\noindent \textbf{Adaptive Digital Agents} \quad
An important thread of agent-improving methods is to build adaptive agents that can autonomously self-improve from experiences. 
Most works focus on integrating past experiences into agent memory by collecting human annotation \citep{deng2024mind2web} or LM-based synthesis \citep{ou2024synatra,xu2025agenttrek}, especially via agent-driven exploration with instruction- \citep{murty2024bagel} or trajectory-driven \citep{murty2024nnetscape} approaches, offering warm starts on the websites of interest. 
Other works gather experiences \citep{wang2024agent} or feedback \citep{qu2024recursive} during test time, and augment them into memory through parametric channels such as supervised fine-tuning \citep{murty2024nnetscape}, contrastive learning \citep{song2024trial}, or reinforcement learning \citep{zhou2024proposer}. 
Meanwhile, non-parametric approaches can directly augment reflections \citep{shinn2023reflexion}, raw past experiences \citep{wang2023voyager,zheng2023synapse}, or further induced reusable workflows \citep{wang2024agent}.
While these adaptive agents learn textual skills stored in memory, our \asi stores skills as verifiable and composable programs in the agent action space (i.e., skill library), thus enabling better quality and efficiency.

\noindent \textbf{Skill Discovery and Learning} \quad
Learning specialized skills for tasks in programmatic \citep{shin2019program,ellis2023dreamcoder,cai2024large,wang2024trove,grand2024lilo}, embodied \citep{sharma2022skill,wang2023voyager,liang2023code,sarch2024vlm,wong2024learning}, and physical \citep{yu2023language} environments has shown to success in agent performance.
Particularly for digital agents built for web navigation tasks, most works focus on exploring skills offline with RL roll-outs \citep{gur2018learning,zheran2018reinforcement,putta2024agent,qi2024webrl} or LM-based prompting \citep{zhou2024proposer,murty2024nnetscape,patel2024large}. While this exploration stage could offer some supervised data to update the agent policy either parametric \citep{murty2024nnetscape,patel2024large} or non-parametrically \citep{zheng2023synapse,murty2024bagel}, it often costs enormous extra computation and may suffer from the lack or mismatch in distribution with the downstream tasks at hand \citep{wang2024agent}.
In contrast, our \asi does not rely on supervised data and can directly learn skills online without prior exploration.

\noindent \textbf{Web Navigation Benchmarks} \quad
Digital agents have been explored across a wide range of tasks \citep{yao2024tau,kapoor2025omniact,xie2024osworld}, among which one of the most popular application being browsing and navigating through versatile websites such as shopping \citep{yao2022webshop}, social media communication \citep{zhou2024webarena,koh2024visualwebarena}, knowledge work tasks \citep{drouin2024workarena}, and more \citep{deng2024mind2web}.
Our work focuses on general web navigation tasks using the WebArena \citep{zhou2024webarena} benchmark, meanwhile exploring other challenging scenarios such as scaled-up activities \citep{yoran2024assistantbench} and cross-domain generalization \citep{deng2024mind2web}.
\section{Conclusion and Future Discussions}

In this work, we present \asi to support web navigation agents to autonomously induce, verify, learn, and apply programmatic skills during online inference. Beyond achieving 23.5\% success rate and 15.3\% efficiency increases in general web tasks, we also showcase \asi's strengths for scaled-up web activities, thanks to the high-level action interface offered by the programmatic abstraction. Moreover, we examine skill generalizability to new, real-world websites, and find \asi still offers great efficiency while flexibly updating skills to new environments.
While our work aims to offer insights on the optimal representation in agent skill acquisition, we still find multiple pieces in \asi worthy of further investigation, such as the conceptually or empirically suitable granularity of skills, the stability of the online evolving process, and the skill quality in comparison to human expert desiderata.


\section*{Acknowledgments}
We would like to thank Jiayuan Mao, Yueqi Song, Boyuan Zheng, and Yu Su for the insightful discussions. We thank Yiqing Xie, Xinran Zhao, and Mingqian Zheng for their helpful comments on the paper draft. Zora is supported by the CMU Presidential Fellowship and Fujitsu Research. Apurva is supported by Amazon.


\bibliography{colm2025_conference}
\bibliographystyle{colm2025_conference}

\clearpage
\appendix
\section{Experiment Details}
\label{app:expr-details}

\subsection{Agent Action Space}
\autoref{tab:action-space} shows the default action space the web navigation agents we employed in all the experiments. This action space remains the same for both (i) static, vanilla agent, as well as the (ii) adaptive agent that learn textual skills in memory, i.e., \awm.

\begin{table}[ht]
\centering
\small
\vspace{-2mm}
\resizebox{0.80\textwidth}{!}{
\begin{tabular}{l|l}
\toprule
\multicolumn{1}{c|}{\bf Action Type} & \multicolumn{1}{c}{\bf Description} \\
\midrule
{\tt noop(wait\_ms)} & {Do nothing for specified time.} \\
{\tt click(elem)} & {Click at an element.} \\
{\tt hover(elem)} & {Hover on an element.} \\
{\tt fill(elem, value)} & {Type into an element.} \\
{\tt keyboard\_press(key\_comb)} & {Press a key combination.} \\
{\tt scroll(x, y)} & {Scroll horizontally or vertically.} \\
{\tt select\_option(elem, options)} & {Select one or multiple options.} \\
\midrule
{\tt goto(url)} & {Navigate to a url.} \\
{\tt go\_back()} & {Navigate to the previous page.} \\
{\tt go\_forward()} & {Navigate to the next page.} \\
\midrule
{\tt new\_tab()} & {Open a new tab.} \\
{\tt tab\_close()} & {Close the current tab.} \\
{\tt tab\_focus(index)} & {Bring tab to front.} \\
\midrule
{\tt send\_msg\_to\_user(text)} & {Send a message to the user.} \\
{\tt report\_infeasible(reason)} & {Notify user that instructions are infeasible.} \\
\bottomrule
\end{tabular}
}
\vspace{-2mm}
\caption{Initial action space for agents throughout our experiments.}
\label{tab:action-space}
\end{table}

\subsection{Prompts for LLM-Based Components}

\begin{tcolorbox}[colback=gray!10, colframe=gray!60, coltitle=black,
                  title=Prompt for Episode Evaluation, fonttitle=\bfseries]
You are an expert in evaluating the performance of a web navigation agent. The agent is designed to help a human user navigate a website to complete a task. Given the user's intent, the agent's action history, the final state of the webpage, and the agent's response to the user, your goal is to decide whether the agent's execution is successful or not. Please be careful of each detail and strict about the evaluation process.

There are three types of tasks:
1. Information seeking: The user wants to obtain certain information from the webpage, such as the information of a product, reviews, map info, comparison of map routes, etc. The bot's response must contain the information the user wants, or explicitly state that the information is not available. Otherwise, e.g. the bot encounters an exception and respond with the error content, the task is considered a failure. Besides, be careful about the sufficiency of the agent's actions. For example, when asked to list the top-searched items in a shop, the agent should order the items by the number of searches, and then return the top items. If the ordering action is missing, the task is likely to fail.
2. Site navigation: The user wants to navigate to a specific page. Carefully examine the bot's action history and the final state of the webpage to determine whether the bot successfully completes the task. No need to consider the bot's response.
3. Content modification: The user wants to modify the content of a webpage or configuration. Carefully examine the bot's action history and the final state of the webpage to determine whether the bot successfully completes the task. No need to consider the bot's response.

*IMPORTANT*
Please be strict about the evaluation process.
Format your response into two lines as shown below:

Thoughts: $<$your thoughts and reasoning process$>$

Status: "success" or "failure"
\end{tcolorbox}

\begin{tcolorbox}[colback=green!5, colframe=green!50!black, coltitle=white, 
                  title=Prompt for Episode Cleaning, fonttitle=\bfseries]
You are a helpful assistant in summarizing web browsing actions. Your task is to summarize the main action taken and the corresponding state change, from the input thought process and environmental description. 

Example input: 
'''Let me think about this step by step:
1) I've clicked the Columns button and based on the accessibility tree, I can now see a list of available columns that can be shown/hidden.
2) Looking through the available columns in the accessibility tree, I don't see any direct column for "Number of Orders" or similar metric.
3) Since I don't see the information I need in the available columns, this suggests that:
   - Either this information is available through a different path
   - Or we need to access the Sales/Orders section to correlate this information
4) The most logical path would be to:
   - Go to the Sales section where we can see all orders
   - And then analyze which customers have the most orders
5) From the accessibility tree, I can see the "SALES" link with bid "183" in the main navigation.

Let me click on the SALES link:
This should open the Sales menu where we can access the Orders section to analyze customer order histories.'''
 
Example output:
'''Clicked the SALES link to access the Orders section for customer order histories.'''
\end{tcolorbox}

\begin{tcolorbox}[colback=orange!10, colframe=red!60!black, coltitle=white, 
                  title=Prompt for Skill Induction, fonttitle=\bfseries]
You are a proficient software engineer. Your task is to (1) summarize reusable functions as APIs from the provided action trajectories, and (2) rewrite the trajecoties using the reusable functions you generated in (1).

For (1), from the provided examples about the same task, you job is to generate Python functions that can be reused to solve (part of) these tasks.
The functions should have mediocre complexity: (i) containing at least three actions and not too simple (e.g., a single line of code), (ii) not too complex (e.g., more than 10 lines of code), and should be general enough to be applied to other similar tasks. The arguments to these functions should be common variables (such as strings and lists), avoid using complex inputs such as another function.

The actions you can use are:
'click', 'fill', 'hover', 'keyboard\_press', 'scroll';
'tab\_focus', 'new\_tab', 'tab\_close', 'go\_back', 'go\_forward';
'send\_msg\_to\_user', 'report\_infeasible', 'select\_option'.
Do not use other undefined actions. Do not include any try-except blocks in the functions.

Please include 'Args', 'Returns', and 'Examples' in the function documentation.

For (2), write the instruction and rewritten code of each example. Do not include the answer response or example-specific information in the rewritten code.
Pay attention to whether all link IDs are available before specifying them in the generated functions.
If you use `send\_msg\_to\_user`, make sure the message is decided within the function, instead of provided as an argument.

Make sure each function contains no less than 2 steps, and no more than 5 steps; to keep the functions simple and task-oriented.
You can generate zero, one, or multiple functions depending on the provided examples.
\end{tcolorbox}

\section{Skill Induction: Analysis}
\label{app:skill-analysis}

We provide more details about the skill induction process, in skill curation and reuse frequency (\S\ref{app:c.1:skill-reuse}) and representative skill case studies (\S\ref{app:c.2:skill-cases})

\subsection{Skill Induction and Reusability}
\label{app:c.1:skill-reuse}
To provide more insights on how agents curate and reuse programmatic skills, for the main experiments on WebArena, we calculate the number of examples that (i) attempt to induce a new skill, (ii) successfully induce a new skill, and (iii) reuse a previously induced skill. 

As shown in \autoref{tab:skill-count}, the agent reuses at least 1 skill for 42.5\% of the cases, which is pretty frequent across websites. Moreover, these skills are created using 7.6\% of the tasks, demonstrating the high scalability of our skill learning approach.

\begin{table}[ht]
\centering
\small
\vspace{-2mm}
\resizebox{0.90\textwidth}{!}{
\begin{tabular}{l|cccc}
\toprule
\multirow{2}{*}{\bf Domain} & \multicolumn{4}{c}{\bf Number of Examples} \\
{} & {\bf Attempted Induction} & {\bf Successful Induction} & {\bf Reuse Examples} & {\bf Total} \\
\midrule
{shopping} & {21} & {8} & {96} & {180} \\
{admin} & {38} & {15} & {108} & {156} \\
{reddit} & {24} & {11} & {14} & {105} \\
{map} & {13} & {10} & {80} & {109} \\
{gitlab} & {25} & {11} & {10} & {175} \\
\bottomrule
\end{tabular}
}
\vspace{-2mm}
\caption{Analyzing the number of examples that attempt, induce, and reuse skills.}
\label{tab:skill-count}
\end{table}

\subsection{Representative Skills}
\label{app:c.2:skill-cases}
We demonstrate two representative types of skills, which (i) chain frequently-used-together actions, and (ii) involve advanced programming primitives.

\paragraph{Chaining Primitive Actions}
The most common type of skills chains multiple frequently-used-together primitive actions (e.g, \texttt{click}, \texttt{fill}) in a certain order, to reflect a particular common usage, such as the \texttt{search\_product} tool illustrated below. This type of skill provides an easy portal for frequent workflows.

\begin{lstlisting}
def search_product(name: str):
    click("Search textbox...")
    if has_popup_window():
        click("Close")
    fill("Search textbox", name)
    click("Search button")
\end{lstlisting}

\paragraph{Advanced Programming Primitives}
Besides a sequential chaining of actions, some skills contain more advanced logics, such as if-else or for/while loops, as the two examples below.

\begin{lstlisting}
def navigate_and_sort_category(category_id: str, subcategory_id: str | None = None):
   """Navigate to a product category and sort items by price.
  
   Args:
       category_id: ID of the main category link
       subcategory_id: Optional ID of the subcategory link, if needed
  
   Returns:
       None
      
   Examples:
       navigate_and_sort_category('1241', '1873')  # PS4 under Video Games
       navigate_and_sort_category('1245')  # For main category only
   """
   click(category_id)  # Click main category
   if subcategory_id:
       click(subcategory_id)  # Click subcategory if provided
   select_option("1553", "Price")  # Sort by price ascending
\end{lstlisting}

\begin{lstlisting}
def browse_category_by_navigation(menu_id_sequence: list):
   """Browse products by navigating through a sequence of menu IDs.
   This function allows navigation through a series of menu interactions.
   Args:
       menu_id_sequence (list): A list of menu IDs to interact sequentially, using hover actions followed by a click.


   Example usage:
       browse_category_by_navigation(['735', '786', '797']) # Navigates Home & Kitchen -> Storage & Organization -> Racks, Shelves & Drawers
   """
   for idx, menu_id in enumerate(menu_id_sequence[:-1]):
       hover(menu_id)
   click(menu_id_sequence[-1])  # Click the final id to land in the predefined category
\end{lstlisting}
\section{Scaled-Up and Cross-Website Tests}
\label{app:other-tasks}

We provide the full list of tasks used in scaled-up (\S\ref{sec:expr-scaled-up}) and cross-website (\S\ref{sec:expr-cross-web}) analyses in \S\ref{app:b.1:scale-up-tasks} and \S\ref{app:b.2:cross-web-tasks}, respectively. In \S\ref{app:b.3:significance-testing}, we further perform significance testing to validate the findings from \autoref{tab:scaled-up-results} and \autoref{tab:crossweb-results}. 

\subsection{Scaled-Up Tasks}
\label{app:b.1:scale-up-tasks}

\autoref{tab:scaled-up-tasks-shopping}, \autoref{tab:scaled-up-tasks-admin}, \autoref{tab:scaled-up-tasks-forum}, \autoref{tab:scaled-up-tasks-gitlab} and \autoref{tab:scaled-up-tasks-map} shows example scaled-up tasks studied on the shopping, admin, social forum, software development, and map websites.

\begin{table}[ht]
\centering
\small
\renewcommand{\arraystretch}{1.3} 
\begin{tabularx}{\textwidth}{X|X|c}
\toprule
\multicolumn{1}{c}{\bf Instruction} & \multicolumn{1}{c}{\bf Checkpoints} & {\bf Score} \\
\midrule
{Add a wireless headphone, a water bottle, a notebook, a ground coffee, and a mug to my shopping cart.} & {Add a wireless headphone to cart; Add a water bottle to cart; Add a notebook to cart; Add a ground coffee to cart; Add a mug to cart.} & {5} \\
\midrule
{Add the most expensive item from the video games category, the cheapest item from the Office Products category, and the most relevant coffee mug to my shopping cart.} & {Add the most expensive item from the video games category to cart; Add the cheapest item from the Office Products category to cart; the most relevant coffee mug to my shopping cart.} & {3} \\
\midrule
{Add the cheapest wireless headphone, a water bottle, the most expensive notebook, a ground coffee, and a mug to my shopping cart.} & {Add the cheapest wireless headphone to cart; Add a water bottle to cart; Add the most expensive notebook to cart; Add a ground coffee to cart; Add a mug to cart.} & {5} \\
\midrule
{Show me the ordered items for each cancelled order from Feb to May in 2023.} & {Show me the 5/17/23 order; Show me the 2/24/23 order; Show me the 2/11/23 order.} & {3} \\
\midrule
{Iterative update my billing address to 231 Willow Way, Suite 100, Chicago, IL, 60601. Then, update my shipping address to 987 Sycamore Circle, Philadelphia, PA, 19102.} & {Successfully update my billing address; Successfully update my shipping address.} & {2} \\
\bottomrule
\end{tabularx}
\vspace{-2mm}
\caption{Exemplar scaled-up browsing tasks on the shopping website.}
\label{tab:scaled-up-tasks-shopping}
\end{table}

\begin{table}[ht]
\centering
\small
\renewcommand{\arraystretch}{1.3} 
\begin{tabularx}{\textwidth}{X|X|c}
\toprule
\multicolumn{1}{c}{\bf Instruction} & \multicolumn{1}{c}{\bf Checkpoints} & {\bf Score} \\
\midrule
{Tell me the the number of reviews that our store received by far that mention terms `disappointed', `satisfied', `decent', `not useful', and `best'.} & {Return the correct number for terms `disappointed', `satisfied', `decent', `not useful', and `best'.} & {5} \\
\midrule
{I need to contact a list of customers. Find the customer name and email with phone number 2058812302, 2137418080, 2065555555, 8015551212, and 555-229-3326.} & {Return the correct name and email information for customers with each of the five phone numbers.} & {5} \\
\midrule
{I will need to update our webpage to create a more energetic vibe. Change the page title of `404 Not Found' to `Bruh bro you clicked the wrong page', the page title of `Enable Cookies' to `Cookie monster coming to your place', the page title of `Home Page' page to `This is the home page!!', the page with title `Privacy Policy' to `No privacy policy is needed is this dystopian world', and lastly, change the page `About Us' to `Secret'.} & {Change the page title correctly for each of the five pages.} & {5} \\
\midrule
{I need to generate a bunch of report to show to the store manager in an hour. Could you help me generate a sales order report for the last month, over the last 45 days, and for Q1? I'll also need a refund report for last year, and a tax report for this year. Today is 3/15/2023.} & {Generate a sales report for 2/1/2023-2/29/2023; generate a sales report for 1/29/2023-3/15/2023; generate a sales report for 1/1/2023-3/15/2023; Generate a refund report for 1/1/2022-12/31/2022; Generate a tax report for 1/1/2023-3/15/2023.} & {5} \\
\midrule
{Tell me the SKU of products that have 10 units, 3 units, and 0 units left. Also, give me the product names that have 2-3 units left.} & {Return the correct SKU for the first three questions; return the correct product names for the last question.} & {4} \\
\bottomrule
\end{tabularx}
\vspace{-2mm}
\caption{Exemplar scaled-up browsing tasks on the shopping admin website.}
\label{tab:scaled-up-tasks-admin}
\end{table}

\begin{table}[ht]
\centering
\small
\renewcommand{\arraystretch}{1.3} 
\begin{tabularx}{\textwidth}{X|X|c}
\toprule
\multicolumn{1}{c}{\bf Instruction} & \multicolumn{1}{c}{\bf Checkpoints} & {\bf Score} \\
\midrule
{I'm planning to organize multiple meetups in the next few months. Help me post notices on virtual meetups for the little women on Apr 10th, for Harry Potter in May 15th, and for Jane Eyre in Jan 30th, in the most suitable forums in PostMill.} & {Post Apr 10th meetup; Post about May 15th meetup; Post Jan 30th meetup. All in book-related forums.} & {3} \\
\midrule
{Could you tell me all forums with names related to computer science?} & {must include: deeplearning (1 pt), MachineLearning (1 pt); optionally (get 1 score if include any): science, askscience, technology.} & {3} \\
\midrule
{Find the most relevant posts about jerseycity, newjersey, and nyc; and tell me how different they are.} & {Correctly find post about jerseycity; Correctly find post about newjersey; Correctly find post about nyc; Answer how different they are.} & {4} \\
\midrule
{Thumbs down the top-2 posts in jerseycity, newjersey, and nyc forums, I don't like them.} & {Thumbs down the top-2 posts in the jerseycity forum; Thumbs down the top-2 posts in the newjersey forum; Thumbs down the top-2 posts in the nyc forum.} & {3} \\
\midrule
{Reply ``Thank you! This is super helpful!'' to three posts about long-distance relationship advice.} & {Reply to three posts with the correct message. Need to be relevant to long-distance relationship advice.} & {3} \\
\bottomrule
\end{tabularx}
\vspace{-2mm}
\caption{Exemplar scaled-up tasks on the Postmill website.}
\label{tab:scaled-up-tasks-forum}
\end{table}

\begin{table}[ht]
\centering
\small
\renewcommand{\arraystretch}{1.3} 
\begin{tabularx}{\textwidth}{X|X|c}
\toprule
\multicolumn{1}{c}{\bf Instruction} & \multicolumn{1}{c}{\bf Checkpoints} & {\bf Score} \\
\midrule
{Display the list of issues in the a11yproject/a11yproject.com repository that have labels related to `help needed', and assign the most recent one to the top contributor of this repository.} & {Display the help-wanted issues; find the top contributor; assign him to the most recent help-needed issue.} & {3} \\
\midrule
{Set up a new, empty repository with the name {\tt agent\_skill\_induction}, and create a MIT license file. Then, invite Abishek and Vinta as collaborators.} & {Create a new repository with given name; Create a MIT license inside; Invite both collaborators.} & {3} \\
\midrule
{Start a private project {\tt web\_agent\_android\_xl} with Android template and add primer, convexegg, abishek as members.} & {Create the repository private and with Android template; Invite all three people as members.} & {2} \\
\midrule
{Add the following users to repo {\tt a11y-webring.club} as developer: [abisubramanya27, lahwaacz], and [yjlou, a11yproject] as maintainer.} & {Add abisubramanya27 and lahwaacz as developers; Add yjlou and a11yproject as maintainers.} & {2} \\
\midrule
{Add the following users [abisubramanya27, lahwaacz, yjlou, a11yproject] to repo {\tt a11y-webring.club}, make sure to assign them different roles.} & {Add abisubramanya27 with role 1; Add lahwaacz with role 2; Add yjlou with role 3; Add a11yproject as role 4. Role 1--4 need to be all different.} & {4} \\
\bottomrule
\end{tabularx}
\vspace{-2mm}
\caption{Exemplar scaled-up tasks on the GitLab website.}
\label{tab:scaled-up-tasks-gitlab}
\end{table}

\begin{table}[ht]
\centering
\small
\renewcommand{\arraystretch}{1.3} 
\begin{tabularx}{\textwidth}{X|X|c}
\toprule
\multicolumn{1}{c}{\bf Instruction} & \multicolumn{1}{c}{\bf Checkpoints} & {\bf Score} \\
\midrule
{Search for the closest restaurants, cafes, parking, and banks to Carnegie Mellon University on the map.} & {Return the closest restaurants; Return the closest cafes; Return the closest parking; Return the closest banks.} & {4} \\
\midrule
{I will need to go to multiple places from Carnegie Mellon University today, including the Univ of Pittsburgh, UPMC shadyside, the Schenley park, and Squirrel Hill. Could you should me the driving route to all those places?} & {Show me driving route from CMU to UPitt; Show me driving route from CMU to UPMC; Show me driving route from CMU to Schenley Park; Show me driving route from CMU to Squirrel Hill.} & {4} \\
\midrule
{Show me the route of driving from CMU to University of Pittsburgh, then walking to the Schenley Park; next, bike to UPMC shadyside, and walk to Squirrel Hill after that.} & {Show me CMU $\rightarrow$ Upitt route by car; Show me Upitt $\rightarrow$ Schenley Park route by foot; Show me Schenley Park $\rightarrow$ UPMC route by bike; Show me UPMC $\rightarrow$ Squirrel Hill route by foot.} & {4} \\
\midrule
{Check if the Univ of Pittsburgh, UPMC shadyside, schenley park, and squirrel hill can be reached within one hour by walking, if departing from Carnegie Mellon University.} & {Return yes to route 1, route 2, route 3, and route 4.} & {4} \\
\midrule
{Tell me the coordinates of Univ of Pittsburgh, UPMC shadyside, schenley park, squirrel hill, and CMU in DD format.} & {Return the coordiates of each of the four places.} & {4} \\
\bottomrule
\end{tabularx}
\vspace{-2mm}
\caption{Exemplar scaled-up tasks on the Map website.}
\label{tab:scaled-up-tasks-map}
\end{table}

\subsection{Cross-Website Tasks}
\label{app:b.2:cross-web-tasks}

\autoref{tab:cross-tasks-shopping}, \autoref{tab:cross-tasks-forum}, and \autoref{tab:cross-tasks-map} lists example tasks to test agent generalization abilities on shopping (OneStopMarket to Target), social forum (Postmill to Reddit), and software development (GitLab to GitHub) domains.

\begin{table}[ht]
\centering
\small
\renewcommand{\arraystretch}{1.3} 
\begin{tabularx}{\textwidth}{X|l|c}
\toprule
\multicolumn{1}{c}{\bf Instruction} & \multicolumn{1}{c}{\bf Checkpoints} & {\bf Score} \\
\midrule
{Show me the options for Canon photo printer?} & {Return the correct search result.} & {1} \\
\midrule
{I have a lot of Nintendo Switch game cards now, help me find the best storage option to fit all 11 cards.} & {Return one valid product.} & {1} \\
\midrule
{What is the price range for beauty products?} & {Return the correct price range.} & {1} \\
\midrule
{Show me products under \$25 for woman shoes} & {Display correct products.} & {1} \\
\midrule
{Show the least expensive shoe storage with a minimum storage capacity of 12 pairs.} & {Display correct products.} & {1} \\
\bottomrule
\end{tabularx}
\vspace{-2mm}
\caption{Exemplar shopping tasks on the target website.}
\label{tab:cross-tasks-shopping}
\end{table}

\begin{table}[ht]
\centering
\small
\renewcommand{\arraystretch}{1.3} 
\begin{tabularx}{\textwidth}{X|X|c}
\toprule
\multicolumn{1}{c}{\bf Instruction} & \multicolumn{1}{c}{\bf Checkpoints} & {\bf Score} \\
\midrule
{Tell me the names of books recommended in the latest five posts in the books forum} & {Find the r/books forum; Find the most recent 5 posts; Give the correct answer.} & {3} \\
\midrule
{Tell me the titles of the 5 most recent posts about little women in the books forum} & {Find the r/books forum; Find little women related posts; Sort the posts by newest.} & {3} \\
\midrule
{What are the recommended products for noise-canceling headphones within a budget of \$200 in r/headphones} & {Find the r/headphones forum; Correctly search with noise-canceling, under \$200 requirements; Return a valid headphone recommendation.} & {3} \\
\midrule
{Find 3 pieces of advices about deal with long-distance relationships in a subreddit for relations.} & {Navigate to a forum about relations; find at least 3 pieces of advice from relevant posts.} & {2} \\
\midrule
{Find if there are any jeep wrangler meetups. If so, when and where?} & {Search in jeep wrangler related forums; Return a valid answer based on the search result.} & {2} \\
\bottomrule
\end{tabularx}
\vspace{-2mm}
\caption{Exemplar social forum tasks on the reddit website.}
\label{tab:cross-tasks-forum}
\end{table}

\begin{table}[ht]
\centering
\small
\renewcommand{\arraystretch}{1.3} 
\begin{tabularx}{\textwidth}{X|X|c}
\toprule
\multicolumn{1}{c}{\bf Instruction} & \multicolumn{1}{c}{\bf Checkpoints} & {\bf Score} \\
\midrule
{Tell me the full address of all international airports that are within a driving distance of 30 miles to Carnegie Mellon University} & {Return Pittsburgh International Airport.} & {1} \\
\midrule
{I will arrive Pittsburgh Airport soon. Provide the name of a Hilton hotel in the vicinity, if available. Then, tell me the the walking distance to the nearest supermarket own by a local company from the hotel.} & {Show me the hotels; Find a nearby supermarket; Show me the walking route from the hotel to the supermarket.} & {3} \\
\midrule
{Show me the walking route from nearby hotels to CMU, Pittsburgh that take at most 5 minutes?} & {Find a hotel that meets the walking time requirement; Show me the walking route.} & {2} \\
\midrule
{I am at CMU Pittsburgh, how long it takes to the nearest USPS postal office with different transportation methods?} & {Return travel time by car, by foot, by bus, and by bike.} & {4} \\
\midrule
{Tell me the coordinates of Carnegie Mellon Cafe in DD format.} & {Return the correct coordinates.} & {1} \\
\bottomrule
\end{tabularx}
\vspace{-2mm}
\caption{Exemplar social forum tasks on the Google Maps website.}
\label{tab:cross-tasks-map}
\end{table}

\subsection{Significance Testing}
\label{app:b.3:significance-testing}

\paragraph{Scaled-Up Tasks}

\begin{wraptable}[9]{r}{0.60\textwidth}
\centering
\small
\resizebox{0.60\textwidth}{!}{
\begin{tabular}{l|cc|cc}
\toprule
\multicolumn{1}{c|}{\multirow{2}{*}{\bf Method Pair}} & \multicolumn{2}{c|}{\bf Success Rate}  & \multicolumn{2}{c}{\bf \# Steps} \\
{} & {t-stat} & {p-value} & {t-stat} & {p-value} \\
\midrule
{\asi vs. \awm} & {-2.3601} & {0.0226} & {2.7664} & {0.0068} \\
{\asi vs. \textsc{Vanilla}} & {-4.0922} & {0.0002} & {2.1983} & {0.0296} \\
\bottomrule
\end{tabular}
}
\vspace{-2mm}
\caption{Results of significance testing on \asi's advantages for scaled-up web tasks.}
\label{tab:scaled-up-sig-test}
\end{wraptable}
We conduct t-tests between (i) \asi and \awm, (ii) \asi and \textsc{Vanilla} agent. From the results in \autoref{tab:scaled-up-sig-test}, we find the advantage of \asi in success rate and efficiency improvements, when comparing to both \awm and \textsc{Vanilla} agents, are statistically significant, as indicated by all t-statistics with absolute values over 2 and p-value below $0.05$.

\paragraph{Cross-Web Tasks}
\begin{wraptable}[9]{r}{0.60\textwidth}
\centering
\small
\vspace{-2mm}
\resizebox{0.60\textwidth}{!}{
\begin{tabular}{l|cc|cc}
\toprule
\multicolumn{1}{c|}{\multirow{2}{*}{\bf Method Pair}} & \multicolumn{2}{c|}{\bf Success Rate}  & \multicolumn{2}{c}{\bf \# Steps} \\
{} & {t-stat} & {p-value} & {t-stat} & {p-value} \\
\midrule
{\asi vs. \awm} & {-1.3980} & {0.1673} & {2.1238} & {0.0378} \\
{\asi vs. \textsc{Vanilla}} & {-3.5984} & {0.0007} & {2.5792} & {0.0125} \\
\bottomrule
\end{tabular}
}
\vspace{-2mm}
\caption{Results of significance testing on \asi's advantages for cross-web tasks.}
\label{tab:cross-web-sig-test}
\end{wraptable}

We conduct similar significance testing on cross-web tasks and report the results in \autoref{tab:cross-web-sig-test}.
While \asi does not significantly outperform \awm in success rate, given the presumably greater flexibility of textual workflows, \asi still exhibits significant advantages on the efficiency side. 
Furthermore, comparing \asi to static \textsc{Vanilla} agents, \asi achieves significant improvements in both success rates and efficiency (i.e., reduced number of steps), suggested by $|t| > 2$ and $p < 0.05$.

\end{document}